\def\year{2020}\relax
\documentclass[letterpaper]{article} 
\usepackage{aaai20}  
\usepackage{times}  
\usepackage{helvet} 
\usepackage{courier}  
\usepackage[hyphens]{url}  
\usepackage{graphicx} 
\usepackage{graphicx}  
\usepackage[table,xcdraw]{xcolor}
\usepackage{times}
\usepackage{latexsym}
\usepackage{amssymb,mathtools}
\usepackage{url}
\usepackage{comment}
\usepackage{graphicx}
\usepackage{paralist}
\usepackage{multirow}
\usepackage{algorithm}
\usepackage{algpseudocode}

\title{Just Ask:\\ An Interactive Learning Framework for Vision and Language Navigation}
\author{%
   Ta-Chung Chi\thanks{Work done while the first author was an intern at Alexa AI.}\\
   Carnegie Mellon University\\
   \texttt{tachungc@andrew.cmu.edu}
   \And
   Mihail Eric \\
   Alexa AI \\
   \texttt{mihaeric@amazon.com} \\
   \AND
   Seokhwan Kim\\
   Alexa AI \\
   \texttt{seokhwk@amazon.com} \\
   \And
   Minmin Shen \\
   Alexa AI \\
   \texttt{shenm@amazon.com} \\
   \And
   Dilek Hakkani-tur \\
   Alexa AI \\
   \texttt{hakkanit@amazon.com} \\
}

\begin{document}

\maketitle

\begin{abstract}
In the vision and language navigation task~\cite{anderson2018vision}, the agent may encounter ambiguous situations that are hard to interpret by just relying on visual information and natural language instructions. We propose an interactive learning framework to endow the agent with the ability to ask for users' help in such situations. As part of this framework, we investigate multiple learning approaches for the agent with different levels of complexity. The simplest model-confusion-based method lets the agent ask questions based on its confusion, relying on the predefined confidence threshold of a next action prediction model. To build on this confusion-based method, the agent is expected to demonstrate more sophisticated reasoning such that it discovers the timing and locations to interact with a human. We achieve this goal using reinforcement learning (RL) with a proposed reward shaping term, which enables the agent to ask questions only when necessary. The success rate can be boosted by at least 15\% with only one question asked on average during the navigation. Furthermore, we show that the RL agent is capable of adjusting dynamically to noisy human responses. Finally, we design a continual learning strategy, which can be viewed as a data augmentation method, for the agent to improve further utilizing its interaction history with a human. We demonstrate the proposed strategy is substantially more realistic and data-efficient compared to previously proposed pre-exploration techniques.
\end{abstract}

\begin{figure}[h!]
    \includegraphics[width=0.45\textwidth]{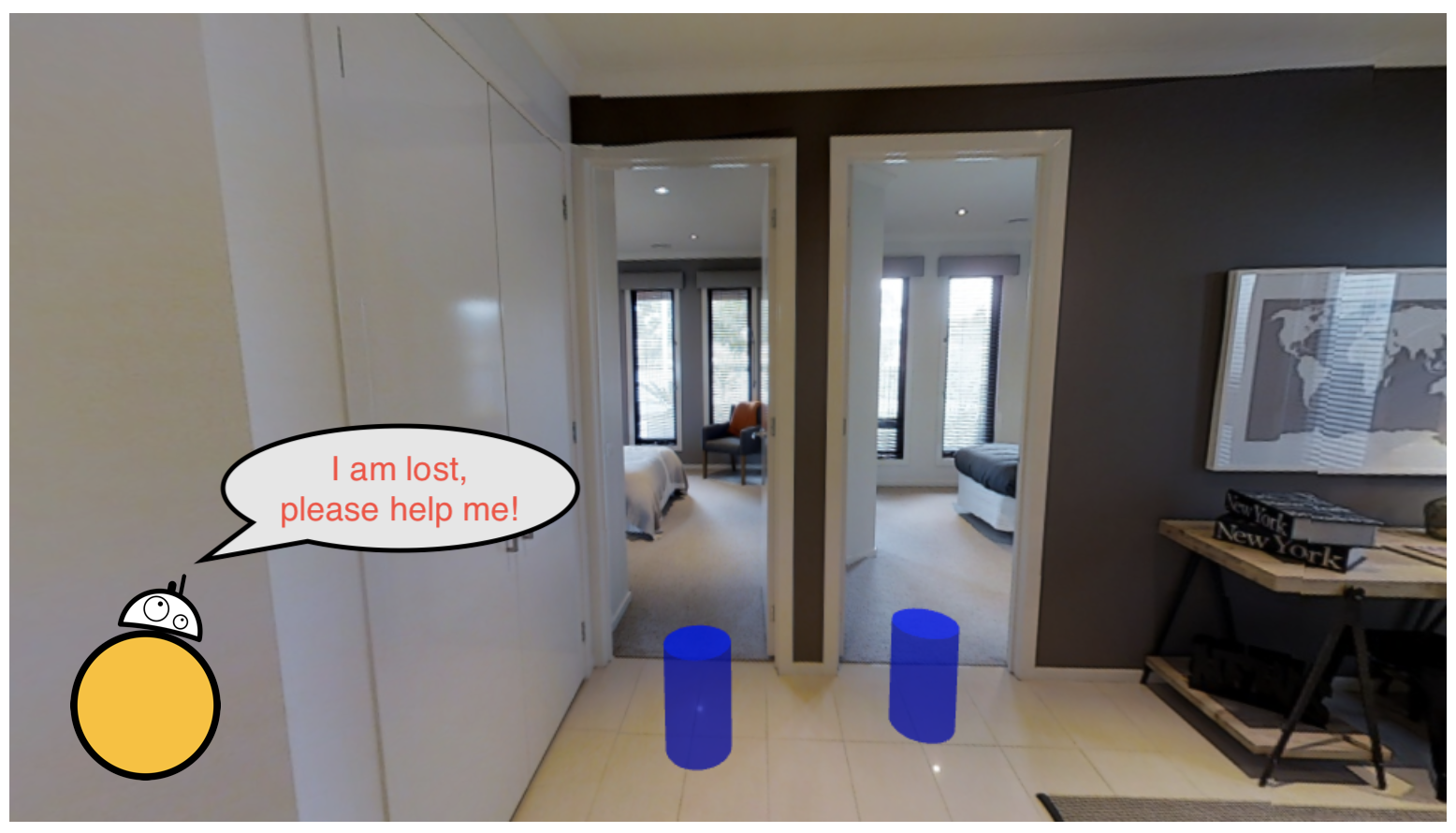}
    \caption{This is a real example from our trained agent. The instruction is~\emph{``...Walk straight, right before you reach the bed."}.  The navigable locations are visualized by blue cylinders. It is impossible to determine which bedroom to enter, and our proposed agent asks for help in this situation.}
        \label{fig:two_room}
        
\end{figure}

\section{Introduction}
Consider the situation in which you want a robot assistant to get your wallet on the bed as in Figure~\ref{fig:two_room} with two doors in the scene and an instruction that only tells it to walk through the doorway. In this situation, it is clearly difficult for the robot to know exactly through which door to enter. 
If, however, the robot is able to discuss the situation with the user, the situational ambiguity can be resolved. For example, the agent can ask the user ``I am confused, please tell me which door to take?" and displays a snapshot on the user's smartphone of what it sees through its camera. The agent can then decide its next action by also considering the user's response.

This scenario suggests that interactive robots can get simple advice from their users to improve completion of tasks, in contrast to their passive counterparts that have no way of getting feedback when problems occur. Indeed, we note that the recent success of virtual assistants can be attributed to their interactive ability with users, demonstrating several human-alike behaviors, such as asking for more information, clarification, and confirmation, which is useful for resolving ambiguities arising naturally in real-world tasks.
Unfortunately, we do not notice such interactive behavior in physical robots. To the best of our knowledge, existing works~\cite{macmahon2006walk,chen2011learning,blukis2018mapping,anderson2018vision,chen2019touchdown} required the robot to complete tasks by itself after the input of preliminary goals and instructions. It has no way to resolve confusions or ambiguities while executing its task, motivating this work's proposed interactive framework. We use the term \emph{robot} and \emph{agent} interchangeably hereafter since our robot lives in a simulator, which can be viewed as a virtual agent.

We propose to extend the Vision and Language Navigation task (VLN) ~\cite{anderson2018vision} that evaluates how well an agent can learn to navigate in an indoor environment and reach a target location by following natural language instructions provided by a human user. To achieve this goal, the agent has to simultaneously understand visual surroundings and natural language.
The inherently ambiguous natural language instructions
and the complexity of the environment may jointly cause confusion and impede the robot's progress.
In addition to the previous two doors example (shown as Figure 1), we observe many vague instructions in the VLN dataset, such as ``walk a bit", where the distance to walk is not clear. In both cases, an agent cannot determine the correct action to execute by only relying on the visual cues and natural language instructions.

Recent approaches to address these difficulties can be mainly characterized into two directions: the first is to explore a better learning framework such that the agent rolls back to previous states when it is confused~\cite{wang2019reinforced,ma2019regretful,ke2019tactical}, and the second is to rely on semi-supervised data augmentation~\cite{fried2018speaker,tan2019learning}. In the first line of research, the length of traversed trajectories tend to be much longer than necessary since the agent has to explore the unknown by itself. On the other hand, the data augmentation approach suffers from several problems including: 1) Previous methods sample a lot of trajectories in the test environment, which would be time and resource inefficient if it were pursued by a robot in real life and 2) these approaches generate augmented data in a user-agnostic way, such that the agent cannot bootstrap off of user-specific patterns. 
With these drawbacks in mind, we ask what a real human would do in such scenarios with insufficient information. The answer is simple -- \emph{just ask for help}.

To investigate the interactive behavior in a principled way, we propose three critical aspects for a learning framework:
\begin{itemize}
    \item \emph{Temporal Resolution} - What is the ideal timing of interactions in the agent action sequence?
    \item \emph{Question Category} - What type of question should be asked? (i.e. request, disambiguation, confirmation, etc.)
    \item \emph{Interaction Form} - How to properly formulate agent questions and human responses for the communication to be efficient?
\end{itemize}
In this work, for~\emph{Temporal Resolution}, the timing is learned either naturally during the navigation or by leveraging human expert knowledge. For~\emph{Question Category}, we focus on the request question, namely ``Which action should the robot take next, amongst the possible next actions?". For~\emph{Interaction Form}, we always use the previous request question, and the human response is the correct next action in the agent's action space. Other types of questions, such as ``Shall I turn left?" (confirmation), ``Shall I turn left or right?" (disambiguation), and how to effectively generate responses in natural language are left as future work.

Two agent models are proposed in this work. The simpler model, called \emph{Model Confusion} or \emph{MC}, mimics human user behavior under confusion. The more complicated model, called \emph{Action Space Augmentation} or \emph{ASA}, is an RL agent with the action space designed specifically to include questions. It automatically learns to ask only necessary questions at the right timing during the navigation thanks to a proposed reward shaping mechanism. To better simulate real-world noise, we design a realistic way to distort answers given by users, while only the high-level RL agent adapts dynamically to different levels of noise.

While the second agent achieves a high success rate, it still struggles with the problem of asking questions in similar situations. To address this concern, we gather the human-agent interaction data, which is used to fine-tune the agent further such that it gets familiar with the environment.

Overall, the main contributions of our work are four-fold:
\begin{itemize}
\item We are among the first to introduce human-agent interaction in the instruction-based navigation task, focusing on successful task completion with minimum questions to users.
\item We propose two interaction methodologies, \emph{MC} and \emph{ASA}, that allow the agent to benefit from human-in-the-loop learning. 
\item We design a simulated user for responding to agent questions and propose alternative ways of creating realistic response data.
\item We use the proposed approach as a data augmentation method, which is useful in a continual learning scenario, such that the agent can improve its performance continually in customers' home.
\end{itemize}

\section{Related Work}
\paragraph{Instruction-Based Navigation}
The instruction-based navigation tasks that use natural language and vision to perform robot navigation have been investigated extensively, including works done in synthetic environments~\cite{macmahon2006walk,chen2011learning,blukis2018mapping}, or agents trained in photo-realistic environments~\cite{anderson2018vision,savva2019habitat,chen2019touchdown}.
The VLN task~\cite{anderson2018vision} has received significant attention recently. Several works in this line designed more powerful agents by using panoramic views~\cite{fried2018speaker}, a better exploration strategy~\cite{ma2019regretful,ke2019tactical,ma2019selfmonitoring}, or generating more diverse environments as training data~\cite{tan2019learning}.~\cite{wang2019reinforced} proposed the cross-modal intrinsic reward for better training. However, due to the lack of interaction ability, the best strategy used by these works for the situation in Figure~\ref{fig:two_room} would only be a random guess.

\paragraph{Human Robot Interaction}
Human robot interaction has long been investigated in the artificial intelligence field~\cite{goodrich2008human}, specifically using dialogue as the interaction format to complete physical tasks~\cite{lopes2000human,spiliotopoulos2001human,fong2003collaboration}.
Recently, an end-to-end pipeline was presented~\cite{thomason2019improving} for translating natural language commands to discrete robot actions. Clarification dialogues are used to improve language parsing and concept grounding. However, the dialogues only take place before the navigation process, ignoring the possibility that confusions may arise throughout the navigation.
Another work~\cite{vdn2019} proposed to integrate human-agent interaction by introducing dialogue behavior into the VLN task. The main contribution is a human-to-human dialogue dataset for the navigation task, where two crowd workers are asked to complete the navigation task by interacting with each other. We foresee that the interaction dynamic between two human might be different from that between a human and an agent, that is, agent questions do not necessarily have to be based on human-human confusions, and instead should be based on the modeling approach used and the confusions of the models.
Furthermore, the ultimate task in our opinion is successful task completion during navigation. Hence our proposed framework mainly focuses on when and what to ask for optimizing task success, with a minimum additional load on the user (i.e. the agent is expected to learn to ask the minimum or strictly necessary number of questions).

\paragraph{Active Learning}
The first line of prior works utilize human feedback in an off-policy (i.e no “ask” action in the action space) manner. For example,~\cite{christiano2017deep} tries to fit the reward function using human comparison on collected trajectories. However, due to its offline training nature, it’s hard to tell how human comparisons can benefit the agent directly during test time.
In contrast, our agent learns the “ask” behavior in an on-policy manner (i.e our “ask” action is in the action space), which means that the agent can still actively ask for help from a human to complete the navigation.
The second line of works rely on some pre-defined metrics and recovery heuristics to ask for human help. For example,~\cite{subramanian2016exploration} encourages the agent to ask for demonstrations when the discrepancy of a state is high.~\cite{knepper2015recovering,tellex2014asking} compare the expected state of the world to the actual state, and if the robot asks for help, a human will repair the failure condition. However, recovery heuristics require human effort for every added argument and do not generalize well. Moreover, these models do not learn the timing for help, which may suffer from the drawback as in our MC agent.

\paragraph{Data Augmentation}
Data augmentation has been shown to be an effective way to further boost the performance of the agent as pointed out in~\cite{fried2018speaker,ma2019selfmonitoring,tan2019learning}. The common approach, borrowing inspiration from unsupervised machine translation~\cite{sennrich2015improving,lample2017unsupervised}, is to sample some trajectories in the environment and use the speaker model~\cite{fried2018speaker} to generate fake natural language instructions.  They benefit from their unsupervised nature since there is no need for human effort. However, the amount of the augmented data is very large, which is only feasible during the training phase assuming that we have a simulator. Others~\cite{wang2019reinforced,tan2019learning} extended this strategy to the unseen/test environment, but the agent still has to explore the environment for a large number of turns, which is too energy-inefficient and slow for the real world. Moreover, the agent cannot learn user-specific characteristics using the generated and fake instructions. In contrast, the human-agent interaction we propose can be used to generate augmented data naturally and efficiently, which only needs human to answer a few questions. Since the instructions are generated by human, they are user-specific by nature. This is particularly useful when the goal is to let a human teach the agent with minimum effort.

\section{Problem Formulation}
Our task is the room-to-room navigation problem~\cite{anderson2018vision}. The agent is given a natural language instruction with $l$ words, $I=\{w_1,w_2\cdots w_i\}_{i=1}^{l}$, where each $w_i$ is a word token. This instruction describes the navigation route $R$, which is represented by a sequence of $n$ viewpoints, $R=\{v_1, v_2\cdots v_j\}_{j=1}^{n}$. We use the terms \emph{location} and \emph{viewpoints} interchangeably hereafter, because the simulator used in this work does not support continuous movement between different viewpoints, so the location of the agent must be on a viewpoint.
The agent starts from the start location $v_1$, going to the target location $v_n$. Since determining when to~\emph{stop} is critical, we denote the final location that the agent decides to stop at as $v_{f}$. 

We formulate this navigation problem as a Markov Decision Process (MDP). At time step $t$, the state of the agent is $s_t$ and the possible action space is $\{a_k^t\}_{k=1}^{m}$.
Formally, the state $s_t$ can be represented by the agent's 36-discretized panoramic view $p^t_i$, and its corresponding horizontal heading $\theta^t_i$ and vertical elevation $\phi^t_i$:
\begin{equation}
\label{eq:feature}
f^t_i = \{Res(p^t_i), sin(\theta^t_i), cos(\theta^t_i), sin(\phi^t_i), cos(\phi^t_i)\},
\end{equation}
where $i$ is between 1 to 36 and $Res(p^t_i)$ is a feature vector derived from a ResNet~\cite{he2016deep} pretrained on imagenet~\cite{imagenet_cvpr09}.

For each of the 36 viewing angles, the environment provides the corresponding next navigable locations. The collection of these locations and a~\emph{stop} action to end the navigation constitute the action space $\{a^t_k\}_{k=1}^m$, where each location is also represented by Eq.~(\ref{eq:feature}). Note that $m$ may not be the same for different time step $t$. This number is determined by the simulator, since the agent may be blocked by obstacles at some of the 36 viewing angles.
Once the agent picks an action $a^t\in \{a^t_k\}_{k=1}^{m}$, the agent moves to a new location with state $s_{t+1}$. To solve the MDP, our baseline model follows the previous idea~\cite{tan2019learning}, which leverages imitation learning and RL techniques.

\subsection{Interaction Ability}
\label{sec:interact_ability}
The agent may encounter ambiguities or get lost during the navigation. Therefore, it is desirable to endow the agent with the ability to interact with a human. Whenever the agent is confused, it sends out a signal~``\emph{I am lost, please help me!}" to the simulated user and asks for help. We assume the simulated user is an oracle $O$, which means it knows where the agent is and returns the next shortest path action to $v_n$.
However, this is only feasible in the simulated environment because real users make mistakes when giving step-wise instructions due to various reasons, including the complex 3D environment.
To simulate this test time error, we assume that users make mistakes with probability $C$. In this case, we calculate the angular differences between the shortest path action $\theta_{sh}$ and the remaining $m-1$ ones, which are then normalized by a softmax function. We do not sample randomly because even when users make mistakes, we assume that they are more likely to provide actions close to the shortest path action. Formally, $softmax_{i\neq sh}(-|\theta_{sh}^t - \theta_i^t|)$.
The distorted answer is sampled from this distribution.

\begin{figure*}
  \centering
  
  \includegraphics[width=0.9\linewidth]{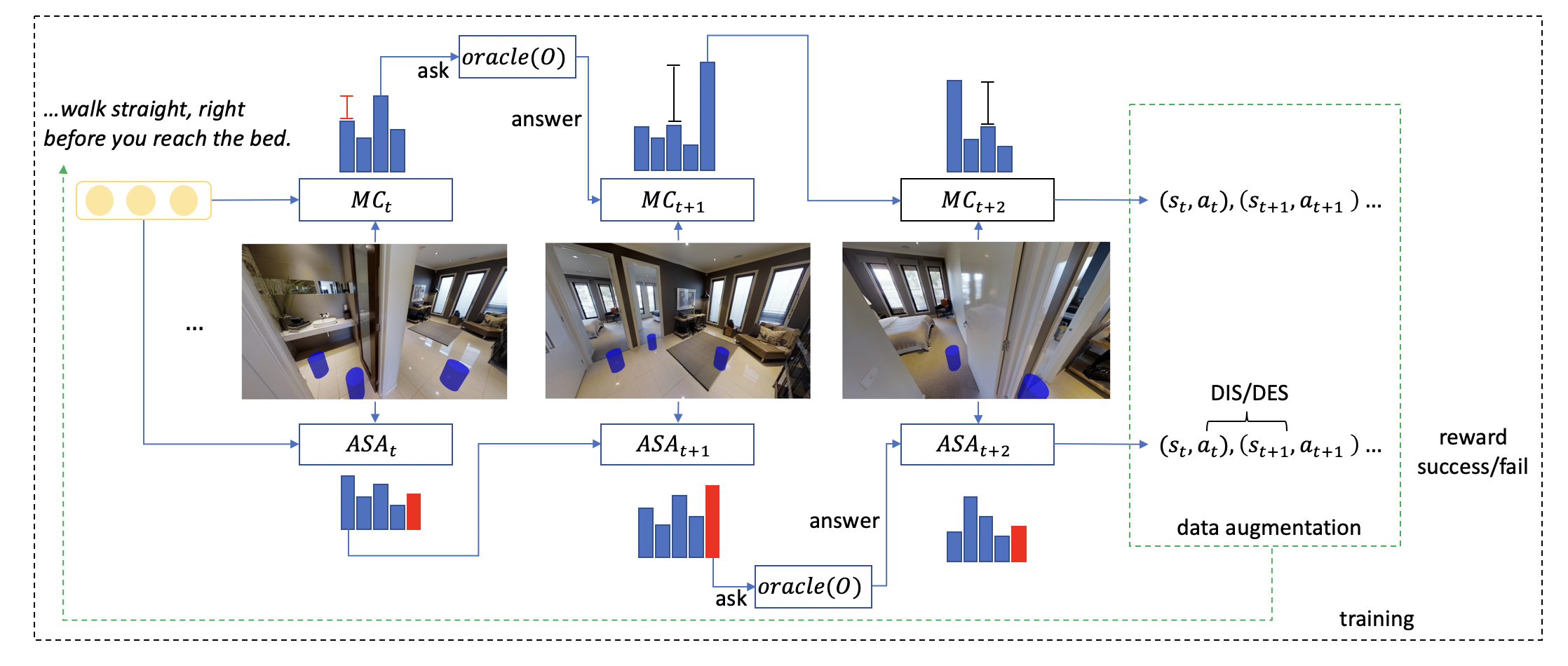}
    
  \caption{A natural language instruction is given in the beginning. As navigation proceeds, the agent decodes the action probability based on actions taken and the current visual clue. The upper one is the~\emph{MC} agent. If the difference of the top2 action probability is less than a threshold, it asks a question and the oracle answers it. The lower one is the~\emph{ASA} agent. There is an additional red bin indicating the~\emph{ask} action. If the agent picks the~\emph{ask}, it sends a signal to the oracle for help. After the agent is trained, we can run it in the unseen environment, collecting trajectories as augmented data.}
  
  \label{fig:model}
\end{figure*}

\section{Model}
\label{sec:Model}
\subsection{Architecture}
Our base model architecture is inspired by previous work in VLN~\cite{fried2018speaker,tan2019learning}.
Each word of the instruction is represented by its word vector:
\begin{equation}
\tilde w_i = Embedding(w_i)
\end{equation}
A bidirectional LSTM model is used to encode the instruction:
\begin{equation}
\{u_i\}_{i=1}^l = LSTM (\{\tilde w_i\}_{i=1}^l)
\end{equation}
where $u_i$ is the concatenation of forward and backward output.
The attentive panoramic view serves as the visual input:
\begin{equation}
    \tilde f^t_i = \sum_{i=1}^{36} \alpha_i^t f^t_i
\end{equation}
with the weight calculated as:
\begin{equation}
    \alpha^t_i = softmax_i(f_i^t W_f \tilde h^{t-1})
\end{equation}
where $\tilde h^{t-1}$ is the previous instruction-aware hidden state, and $W_f$ is a learnable matrix.

The decoder is an auto-regressive LSTM, where the input of every time step is the concatenation of the previous action and the attentive visual feature:
\begin{equation}
    h^t = LSTM([\tilde f^t;a^{t-1}], \tilde h^{t-1})
\end{equation}
It is desirable that the agent focus on the right part of the instruction throughout the navigation process. Hence we calculate the attentive instruction:
\begin{equation}
    \tilde u^t = \sum_i ^l\beta^t_i u_i
\end{equation}
The weight is calculated over all words of instruction:
\begin{equation}
    \beta^t_i = softmax_i(u_i W_u h^t)
\end{equation}
$W_u$ is a learnable matrix.
The hidden state is calculated as:
\begin{equation}
    \tilde h^t = tanh(W_h(\tilde u^t; h^t))
\end{equation}
This instruction-aware hidden state is passed to next time step $t+1$ and used for computing the action distribution at current time step $t$.

\begin{equation}
    P^t(\{a^t_k\}_{k=1}^m) = softmax_k(g^t_k W_a \tilde h^t)
    \label{eq:action_p}
\end{equation}
$g^t_k$ is calculated as the same as Eq.~(\ref{eq:feature}) but on the next navigable locations.

\subsection{Optimization}
Whenever the agent goes to a viewpoint $v_t$ at time step $t$, the environment computes $\tilde R_t = \{v_t, v_{t+1} \cdots v_n\}$, which is the shortest path from $v_t$ to the target location $v_n$.
The action going from $v_t$ to $v_{t+1}$ is returned as the teacher action for supervision signal. For the supervised learning loss, we calculate the cross entropy between the teacher action and $P^t$. For RL, the agent samples its actual action from the action distribution $P^t$ computed in Eq.~(\ref{eq:action_p}).
As in previous work, a +2 reward is given if the final location is within 3 meters distance to $v_n$. Otherwise, the reward given is -2. We use the advantage actor critic (A2C)~\cite{konda2000actor} as our RL algorithm.

\section{Methodologies}
We introduce two agents as in Figure \ref{fig:model} of different levels for the human-agent interaction along with the data augmentation strategy that makes use of the interaction data.
\subsection{Training}

\subsubsection{Model Confusion (MC)}
\label{sec:model_confusion}
In our \emph{MC} model, the idea is that if the agent is confident of itself, then the predicted action distribution should be sharp. To quantify the confusion intuition, we first sort action probabilities in a decreasing order, and say an agent is \emph{confused} if the difference of the top two actions is less than a threshold $\epsilon$:
\begin{equation}
p_{sorted}^t[0] - p_{sorted}^t[1] < \epsilon,
\label{eq:top2}
\end{equation}
we provide the agent with the shortest path action. The threshold is used to control the degree of confusion.

In this simple mode, since the timing of whether to ask questions is not trained, we use the original action space without~\emph{ask}. Note that this method can be applied directly on pre-trained models described in sec.~\ref{sec:Model} during test time.

\subsubsection{Action Space Augmentation (ASA)}
\label{sec:ASA}
We introduce as many actions as the types of questions the agent asks in addition to the original action space. In this work, the action space is enlarged by 1 to represent the "What should I do next?" question, which is used to indicate whether to ask for help. Formally, the new action space is $\{a^t_k\}_{k=1}^m \cup ask$, where $ask$ is the question indicator. If the agent chooses the $ask$ action, it will remain in the same state $s_t$ and $O$ will give it the action on the shortest path route to $v_n$. Each selected $ask$ is associated with a negative reward $r_{ask}$,  such that $r_{ask} < 0$ to ensure only necessary questions are asked.
The action probability becomes:
\begin{equation}
    p^t(\{a^t_k\}_{k=1}^{m+1}) = softmax_{k\cup \emph{ask}}(g^t_k W_a \tilde h^t \cup \emph{ask}W_a\tilde h^t)
    \label{eq:action_p_ask}
\end{equation}
We represent the action feature of~\emph{ask} by a vector of dimension same as $g^t_k$  consisting of all ones.

Enlarging the action space alone does not provide the desired question-asking behavior without the help from reward shaping~\cite{ng1999policy}, which is a useful technique in training RL algorithms.
The difference of distance to the goal location between two consecutive steps can be used as the shaping reward~\cite{wang2019reinforced}. However, this will encourage the agent to bias toward the shortest path, but not follow the path indicated by the instruction. We propose to use an additional reward shaping term, the~\textbf{deviation shaping}, to encourage the agent to follow the instruction. We call the original shaping reward as the distance shaping:
\begin{equation}
    DIS_t = d(v_t, v_n)-d(v_{t+1}, v_n)
\end{equation}
where $d(v_t, v_n)$ is the distance between the current location and the target location.
The proposed deviation shaping is:
\begin{equation}
    DEV_t = -d(v_{t+1}, v_{gt})
\end{equation}
$v_{gt}$ is the viewpoint with the shortest distance to $v_t$ among the whole ground truth trajectory. It is straightforward to see that if the agent follows the ground truth trajectory better, the reward should be higher. Moreover, this shaping reward helps reduce the number of questions asked while preserving the same success rate. The reason is better alignment with ground truth trajectories leads to less ambiguities during the navigation. Without~\emph{DEV}, the agent will ask questions~\emph{at every timestep}. These two shaping rewards are summed together during the RL training. Concretely, the critic $CR$ in the A2C algorithm is optimized at every timestep $t$ as:
\begin{equation}
\label{eq:critic_obj}
    (CR_t - (DEV_t + DIS_t + r_{ask}+G_t))^2,
\end{equation}
where $G_t$ is the discounted cumulative reward estimated by the monte carlo method~\cite{sutton1998introduction}.
We note that the~\emph{intrinsic reward} in~\cite{wang2019reinforced} shares a similar idea with our work. However, they train a sequence-to-sequence critic where the input is traversed trajectories and output is instruction decoding probabilities. This critic is used to calculate the cycle reconstruction reward at every timestep, which is much slower than our simple but effective deviation shaping.

\begin{figure}[h!]
    \includegraphics[width=0.45\textwidth]{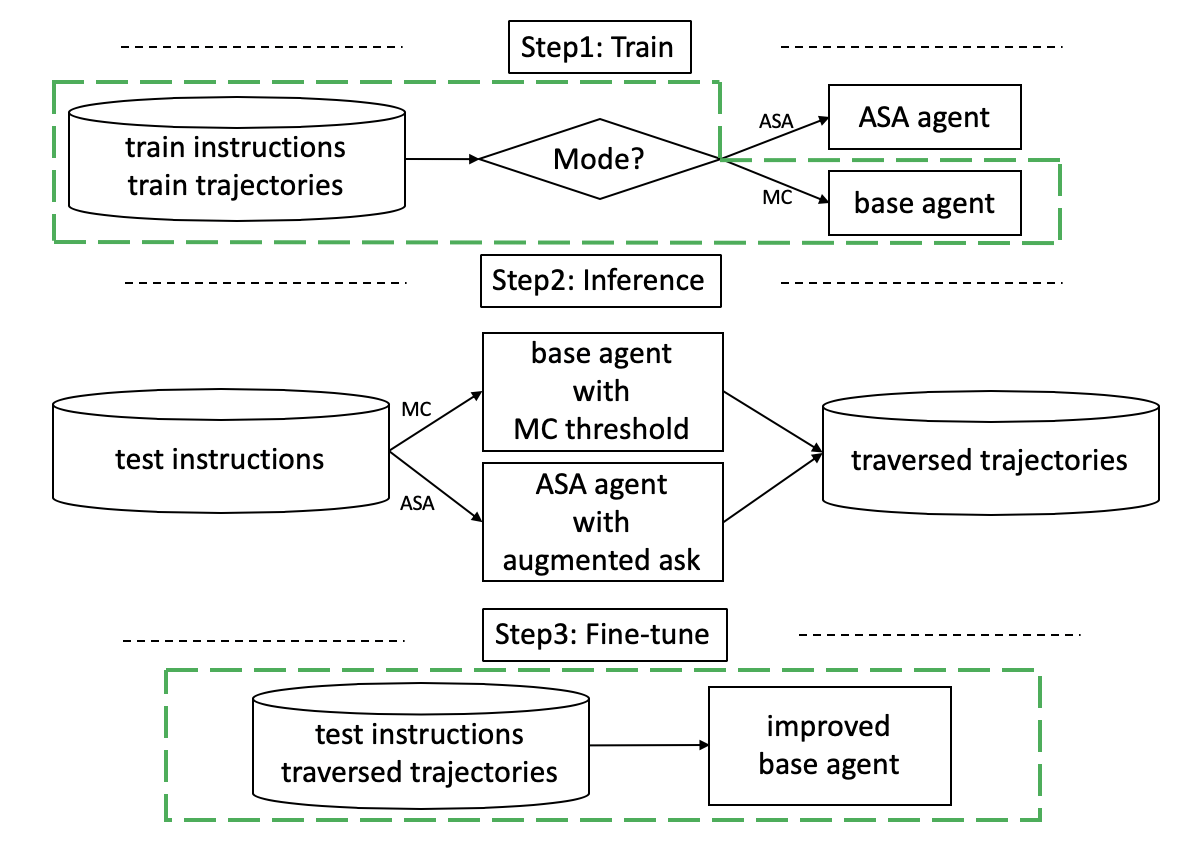}
    
    \caption{This is the data augmentation process. The green dashed boxes indicate the training procedure in~\cite{tan2019learning}. Agents are not allowed to ask questions at step 3. Finally, generalizability is tested with different trajectories and instructions ($T_b$) from the input ($T_a$) of step 3. $T_{a,b}$ are described in sec.~\ref{sec:hge}.}
        \label{fig:algo}
\end{figure}

\begin{table*}[h!]
\centering
\small
\begin{tabular}{cccccccccc}
\hline
\multicolumn{2}{c}{\multirow{2}{*}{Agent Types}} & \multicolumn{2}{c}{success rate} & \multicolumn{2}{c}{number of questions} & \multicolumn{2}{c}{move steps} & \multicolumn{2}{c}{ask percentage} \\ \cline{3-10} 
\multicolumn{2}{c}{}                                                                                  & val\_seen      & val\_unseen     & val\_seen         & val\_unseen         & val\_seen     & val\_unseen    & val\_seen       & val\_unseen      \\ \hline
\multicolumn{2}{c}{Base Model w/o Interaction}                                                                       & 0.551          & 0.471            & -                & -                 & 5.09         & 4.95           & -             & -               \\

\hline\hline
\multirow{5}{*}{MC}
&$\epsilon=0.1$     & 0.614                          & 0.562       & 0.58              & 0.56               & 5.01         & 4.88           & 0.103           & 0.103             \\ \cline{2-10}
&0.2     & 0.693                          & 0.633       & 1.00              & 0.96               & 5.01         & 4.88          & 0.167           & 0.164             \\ \cline{2-10}
&0.3     & 0.759                          & 0.695       & 1.34              & 1.33               & 5.03         & 4.88           & 0.210           & 0.214             \\ \cline{2-10}
&0.4     & 0.822                          & 0.756       & 1.67              & 1.68               & 5.05         & 4.90           & 0.248           & 0.255             \\ \cline{2-10}
&0.5     & \bf 0.854                          & \bf 0.807       & 1.95              & 1.99               & 5.05         & 4.92           & 0.278           & 0.287             \\ \cline{2-10}

\hline\hline
\multirow{5}{*}{ASA}
&$r_{ask}=0.1$     & \bf 0.790          & \bf 0.732            & 1.25              & 1.92               & 5.39         & 5.49           & 0.188           & 0.259             \\ \cline{2-10}
&0.2     & 0.690          & 0.710            & 0.84              & 1.59                & 5.58         & 5.72           & 0.131           & 0.218             \\ \cline{2-10}
&0.3     & 0.704          & 0.676            & 0.74              & 1.41                & 5.36         & 5.42           & 0.120           & 0.206             \\ \cline{2-10}
&0.4     & 0.632          & 0.598            & 0.38              & 0.85               & 5.29         & 5.25           & 0.067           & 0.139             \\ \cline{2-10}
&0.5     & 0.590          & 0.494            & 0.06              & 0.15               & 5.04         & 4.85           & 0.012           & 0.030             \\ 
\hline\hline
\multirow{2}{*}{\shortstack[c]{Human-Guided\\ Exploration}}
&disjoint\dag                                                                     & \bf 0.555          & \bf 0.554            & -                & -                 & 5.19         & 4.91           & -             & -               \\ \cline{2-10}
&random                                                                       & \bf 0.565          & \bf 0.649            & -                & -                 & 5.18         & 4.93           & -             & -               \\ \hline

\multicolumn{2}{c}{Pre-Exploration\dag}                                                                       & 0.560          & 0.504            & -                & -                 & 5.21         & 5.11           & -             & -               \\ \hline
\end{tabular}
\caption{Interactive agents,~\emph{MC} and~\emph{ASA}, both outperform the baseline without interaction. When $\epsilon$ is larger,~\emph{MC} tends to ask more questions while~\emph{ASA} asks fewer questions when $r_{ask}$ is lower. The move steps of~\emph{MC} remain the same with different $\epsilon$, but it increases when $r_{ask}$ of~\emph{ASA} decreases. This is probably because~\emph{ASA} learns the~\emph{ask} behavior during training, so it tries to explore more to maximize the reward. Finally, the proposed human-guided exploration outperforms the pre-exploration techinque~\cite{tan2019learning} by 5\%.~\dag We use the augmented data with size 1500 for a fair comparison.}
\label{tab:all}

\end{table*}

\subsection{Data Augmentation}
\label{sec:data_aug}
Our proposed interaction methods can be used to generate augmented data. Concretely, we execute the trained agent in test environment such that the agent may ask questions and receive answers from $O$. The advantage of doing so is that by answering a few simple questions, the originally wrong trajectories might be corrected. These corrected trajectories serve as augmented data to prevent the agent from making same mistakes. The complete procedure is outlined in Figure~\ref{fig:algo}. As long as users keep using the agent, we can collect more interaction data to fine-tune the agent in a continual learning scenario. 
The differences between our human-guided exploration, and the pre-exploration approach~\cite{wang2019reinforced,tan2019learning} are highlighted in Table~\ref{tab:diff}.

\begin{table}[h!]
\small

\begin{tabular}{cccll}

\cline{1-3}
\multicolumn{1}{|c|}{}             & \multicolumn{1}{c|}{\bf human-guided exploration}          & \multicolumn{1}{c|}{\bf pre-exploration}     &  &  \\ \cline{1-3}
\multicolumn{1}{|c|}{\bf instructions} & \multicolumn{1}{c|}{real, user-specific}     & \multicolumn{1}{c|}{fake, user-agnostic} &  &  \\ \cline{1-3}
\multicolumn{1}{|c|}{\bf trajectories} & \multicolumn{1}{c|}{not shortest, traversed} & \multicolumn{1}{c|}{shortest, sampled}   &  &  \\ \cline{1-3}
\multicolumn{1}{l}{}               & \multicolumn{1}{l}{}                         & \multicolumn{1}{l}{}                     &  & 
\end{tabular}

\caption{The differences between the human-guided exploration versus pre-exploration.}

\label{tab:diff}
\end{table}

\section{Experiments and Discussions}
We describe the R2R dataset used and the performance of our model. We propose an additional evaluation metric to measure the effectiveness of the interactive behavior in the VLN task. The impact of the imperfect oracle is also investigated. Finally, we compare our data augmentation method with previous work in terms of data efficiency.

\subsection{R2R Data Statistics}
\label{sec:data_stat}
In the R2R dataset~\cite{anderson2018vision}, annotators are given image sequences of sampled shortest path trajectories, then they write down natural language instructions that best describe the paths. The dataset contains 21,567 navigation instructions with an average length of 29 words.  The instruction vocabulary consists of around 3100 words due to the nature of the navigation task. The train set includes 61
scenes, with instructions split 14,025 train / 1,020 val seen. 11 scenes and 2,349 instructions are reserved for validating in unseen environments (unseen validation).

\begin{table}[]
\centering
\small
\begin{tabular}{|c|c|c|c|c|}
\hline
$C$ & \multicolumn{2}{c|}{success rate} & \multicolumn{2}{c|}{number of questions} \\ \hline
                    & ASA-0.1          & MC-0.5         & ASA-0.1                & MC-0.5                \\ \hline
0.0                 & 0.732           & 0.807         & 1.92                   & 1.99                  \\ \hline
0.1                 & 0.731           & 0.743         & 2.25                   & 1.99                  \\ \hline
0.2                 & 0.731           & 0.688         & 2.69                   & 1.99                  \\ \hline
0.3                 & 0.732           & 0.619         & 3.55                   & 1.99                  \\ \hline
0.4                 & 0.722            & 0.562         & 4.44                   & 1.94                  \\ \hline
\end{tabular}
\caption{The impact of different noise levels. The 0.1 after ASA indicates the $r_{ask}=0.1$ in Eq.~(\ref{eq:critic_obj}), and the 0.5 after MC indicates the $\epsilon=0.5$ in Eq.~(\ref{eq:top2}).}

\label{tab:noise}
\end{table}

 
\subsection{Evaluation Metric}
\label{sec:evaluation_metric}
We evaluate our agent on the success rate and the number of steps taken which are the standard reported metrics~\cite{anderson2018vision} of the VLN task. An episode is a success if the
navigation error is less than 3 meters. In addition to the standard metrics, we think it is necessary to propose a new evaluation metric to justify the effectiveness of the human-agent interaction, which is the percentage of total actions taken that are \emph{ask} actions: $\frac{\#ask}{\#ask+\#move}$
where $\#{move}$ is the number of moving actions other than $\#ask$.

\begin{figure}
  \centering
  \includegraphics[width=0.85\linewidth]{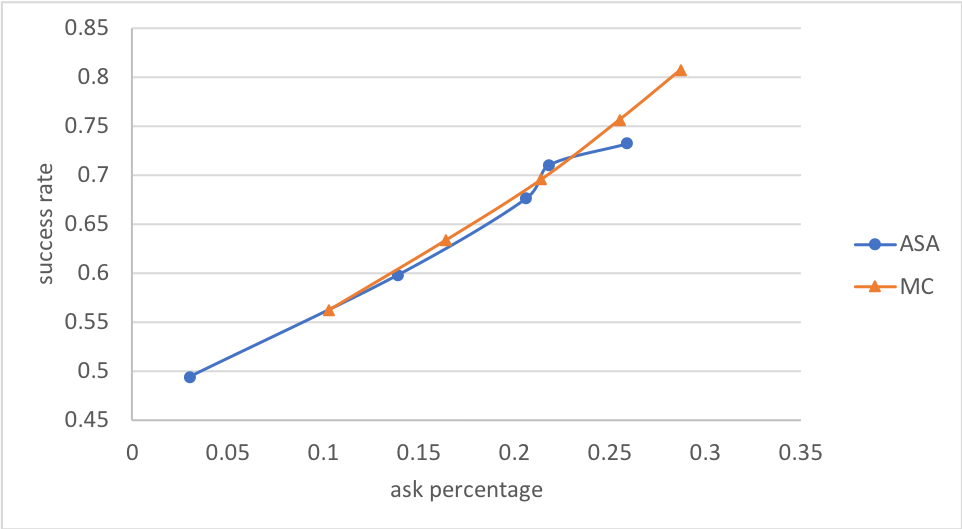}
    
  \caption{The performance of different agents. We can only control the threshold or penalty during training, so the points of the two curves would not be plotted on the same x values.}
  
  \label{fig:ask_diff}
\end{figure}

\subsection{Interaction Results}
\label{sec:asa_mc_res}
In this setting, the agent is allowed to ask questions during test time with an oracle providing shortest path actions. For the~\emph{ASA} agent, we vary the penalty associated, $r_{ask}$, to each~\emph{ask} action. For the~\emph{MC} agent, we adjust the confusion threshold in Eq.~(\ref{eq:top2}). Then we test the agent on the unseen validation split. The results are in Table~\ref{tab:all}.

For the~\emph{MC} agent, the success rate and ask percentage become higher when we increase the threshold $\epsilon$.
The same observation applies to the~\emph{ASA} agent. Note that with a lower penalty $r_{ask}$, the agent is encouraged to ask more questions. At the same time, the success rate and the ask percentage both increase.
If we compare the two agents, the performance is roughly the same regarding the success rate at the same ratio of the ask actions as shown in Figure~\ref{fig:ask_diff}. It is interesting to note that the move steps of the~\emph{ASA} agent increases while the~\emph{MC} agent remains the same. We hypothesize this is because~\emph{ASA} learns the~\emph{ask} behavior during training, so it tries to explore more to maximize the reward. As for~\emph{MC}, the threshold is applied only at test time, it does not learn the exploration behavior. While~\emph{MC} seems to be simpler and more effective than~\emph{ASA}, the~\emph{ASA} agent can adapt to errors in human-agent interactions more easily as we will see.

\subsubsection{Imperfect Oracle}
\label{sec:imperfect_oracle}
We adjust the distortion probability $C$ in sec.~\ref{sec:interact_ability} to see how our agents adapt to different levels of noise. The results are in Table~\ref{tab:noise}. The~\emph{ASA} agent asks more questions with the same success rate while the~\emph{MC} agent asks the same number of questions but the success rate drops linearly. The behavior of~\emph{ASA} is more ideal, since it can adjust dynamically to different levels of noise with the same success rate, which is particularly useful if the agent is a real product.

\subsection{Human-Guided Exploration}
\label{sec:hge}
It is desirable that the agent can further improve its navigation ability after several rounds of interactions. We split the unseen validation data to $T_a$ and $T_b$. This is testing the real use case when a user brings the agent to a new environment ($T_a$) and teaches it through interaction for a while. After that, the agent is evaluated in the same environment with different instructions and paths ($T_b$) to see the effectiveness.
 
\subsubsection{Disjoint Split}
In this setting, $T_a$ and $T_b$ do not share the same trajectories and instructions but the house plans are shared. The reason for this splitting strategy is to compare fairly with the pre-exploration technique since~\cite{tan2019learning,wang2019reinforced} ensured the augmented paths are different from those in the test environment. We run the same experiment with~\emph{ASA} and~\emph{MC} agents on $T_a$ and obtain the interaction history data, which is used to fine-tune the agent. Finally, we test the re-trained agent on $T_b$ without human-agent interaction. There are 2349 instructions in the unseen environment. We use the first 1500 as $T_a$ and the remaining 849 as $T_b$.

\subsubsection{Random Split}
\label{sec:rand_split}
In this setting, $T_a$ and $T_b$ may share the same trajectories but with different instructions. The motivation is to mimic the real-world scenario where customers buy the robot and put it in their houses. It is natural for a human to use different sentences to express the same goal. We randomly permute the unseen validation split and use the first 1500 as $T_a$ and the remaining 849 as $T_b$.

The results of two settings are in Table~\ref{tab:all}. We use the best~\emph{ASA} ($r_{ask}=0.1$) agent to generate augmented data. Better performance is observed on the random split setting because the agent has already seen some trajectories in $T_a$. As for the disjoint setting, we can compare the results of~\emph{ASA} agent fairly with~\cite{tan2019learning}. With the same amount of augmented data (1500 on $T_a$), our method outperforms theirs by 5\% (0.554 vs. 0.504).

Finally, we limit the fine-tune stage (stage 3 in Figure~\ref{fig:algo}) to only supervised training instead of the mixture of supervised and RL training. This is to further reduce the time and energy consumption in test environment which in real-life would be a new customer's home. The result is 0.514 vs. 0.483. While the performance of both methods drops due to the lack of exploration, ours still outperforms the baseline by 3\%.

\subsubsection{Data Efficiency}
\begin{figure}
  \centering
  \includegraphics[width=0.9\linewidth]{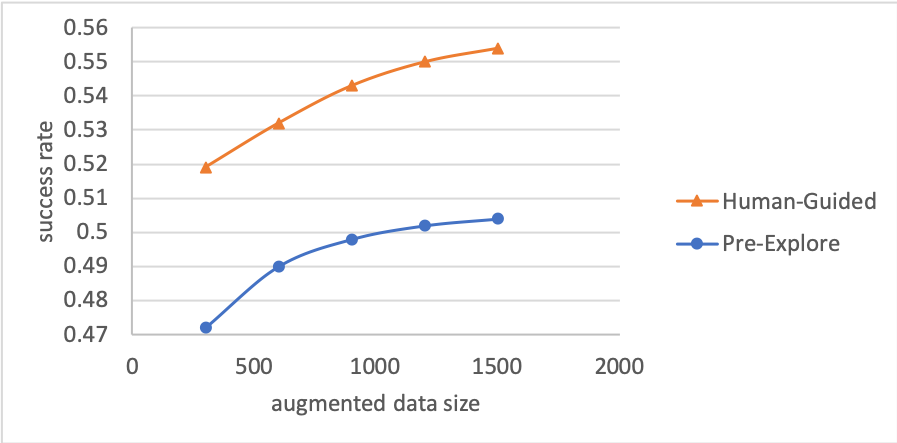}
    
  \caption{The curves of different data augmentation methods. Human-Guided Exploration consistently outperforms Pre-Exploration by a large margin.}
  
  \label{fig:aug_diff}
\end{figure}

We vary the augmented data size of the pre-exploration approach to see its data efficiency.
The results are in Figure~\ref{fig:aug_diff}.
It is easy to see that human-guided exploration can reach the same performance by using much less data, demonstrating that our agent is more data-efficient. Moreover, this experiment highlights the importance of real instructions and trajectories.

\section{Conclusion}
In this paper, we propose an interactive learning framework to make the agent capable of resolving ambiguous situations by interacting with a human during learning or execution time. Two approaches are proposed, model confusion-based (MC) and RL with reward shaping (ASA). Experiment results demonstrate our agents can strike a balance between the task success rate and number of questions asked. Moreover, the RL agent can adapt dynamically to noise. Finally, we propose a strategy to fine-tune the agent using augmented data collected from human-agent interactions, which is more data-efficient and realistic than the previous method.

\bibliography{aaai20}
\bibliographystyle{aaai20}
\end{document}